\documentclass[sigconf]{acmart}
\AtBeginDocument{%
  }

\copyrightyear{2026}
\acmYear{2026}
\setcopyright{cc}
\setcctype{by}
\acmConference[MM '26]{Proceedings of the 34th ACM International Conference on Multimedia}{November 10--14, 2026}{Rio de Janeiro, Brazil.}
\acmBooktitle{Proceedings of the 34th ACM International Conference on Multimedia (MM '26), November 10--14, 2026, Rio de Janeiro, Brazil}
\acmISBN{979-8-4007-2213-4/2026/11}
\acmDOI{10.1145/3767308.3836604}
\usepackage{subcaption}

\makeatletter
\newcommand{\startfourauthorrow}{%
  \g@addto@macro\addresses{%
    \and\par
    \author@bx@wd=\textwidth\relax
    \advance\author@bx@wd by -\author@bx@sep\relax
    \divide\author@bx@wd by 4\relax
    \advance\author@bx@wd by -\author@bx@sep\relax
    \advance\author@bx@wd by -.8em\relax
    \noindent\leavevmode\hspace*{-1em}%
    \let\and\relax}}
\makeatother

\begin{document}

\title{DishSeg24k: A Large-Scale Benchmark for Food Segmentation with Stochastic Expert Decoding}

\author{Yilin Wang}
\affiliation[obeypunctuation=true]{%
  \institution{Institute of Computing Technology, Chinese Academy of Sciences},
  \city{Beijing}, \country{China}
}
\affiliation[obeypunctuation=true]{%
  \institution{University of Chinese Academy of Sciences},
  \city{Beijing}, \country{China}
}
\email{wangyilin241@mails.ucas.ac.cn}

\author{Haochen Shi}
\affiliation{%
  \institution{Hangzhou Dianzi University}
  \city{Hangzhou}
  \country{China}
}
\email{shihc@hdu.edu.cn}

\author{Guanyu Chen}
\affiliation{%
  \institution{North China Electric Power University}
  \city{Beijing}
  \country{China}
}
\email{120252327037@ncepu.edu.cn}

\startfourauthorrow

\author{Weiqing Min}
\correspondingauthor
\affiliation[obeypunctuation=true]{%
  \institution{Institute of Computing Technology, Chinese Academy of Sciences},
  \city{Beijing}, \country{China}
}
\affiliation[obeypunctuation=true]{%
  \institution{University of Chinese Academy of Sciences},
  \city{Beijing}, \country{China}
}
\affiliation[obeypunctuation=true]{%
  \institution{Institute of Intelligent Computing Technology, Chinese Academy of Sciences},
  \city{Suzhou}, \country{China}
}
\email{minweiqing@ict.ac.cn}

\author{Jinkai Zheng}
\affiliation{%
  \institution{Hangzhou Dianzi University}
  \city{Hangzhou}
  \country{China}
}
\affiliation{%
  \institution{Lishui Institute of Hangzhou Dianzi University}
  \city{Lishui}
  \country{China}
}
\email{zhengjinkai3@hdu.edu.cn}

\author{Chenggang Yan}
\affiliation{%
  \institution{Hangzhou Dianzi University}
  \city{Hangzhou}
  \country{China}
}
\email{cgyan@hdu.edu.cn}

\author{Shuqiang Jiang}
\affiliation[obeypunctuation=true]{%
  \institution{Institute of Computing Technology, Chinese Academy of Sciences},
  \city{Beijing}, \country{China}
}
\affiliation[obeypunctuation=true]{%
  \institution{University of Chinese Academy of Sciences},
  \city{Beijing}, \country{China}
}
\affiliation[obeypunctuation=true]{%
  \institution{Institute of Intelligent Computing Technology, Chinese Academy of Sciences},
  \city{Suzhou}, \country{China}
}
\email{sqjiang@ict.ac.cn}

\renewcommand{\shortauthors}{Yilin Wang et al.}

\begin{abstract}
Food segmentation is essential for applications such as intelligent catering, dietary assessment, and food recommendation. However, existing benchmarks do not faithfully capture the dense inter-dish overlap, fine-grained class similarity, and extreme long-tail distributions of real-world dining scenes. To fill this gap, we introduce DishSeg24k, a large-scale dish-level segmentation benchmark with 24,096 images, 112,281 instances, and 278 fine-grained categories in real-world dining environments. Based on DishSeg24k, we further propose Food Expert-Adaptive Segmentation Transformers (FEAST) to address these challenges. FEAST models query-based decoding as a Markov Decision Process (MDP), treating each decoder-layer update as a sequential decision that enables exploration of uncertainty along dish boundaries. We further redesign the decoder with a reinforcement learning (RL)-guided Mixture-of-Experts (MoE) module, in which a decoupled dual-critic optimization scheme separates task-oriented query refinement from structure-aware expert routing. This design promotes expert specialization and prevents expert collapse under long-tail category distributions.
Finally, extensive experiments on DishSeg24k demonstrate that FEAST outperforms the strongest baseline by 3.21\% mIoU, 3.68\% mDice, and 4.00\% mAcc. We further validate the effectiveness of FEAST on FoodSeg103. 
\end{abstract}


 


\maketitle
\section{Introduction}\par
\begin{figure*}[t!]
    \centering
    \includegraphics[width=0.9\linewidth]{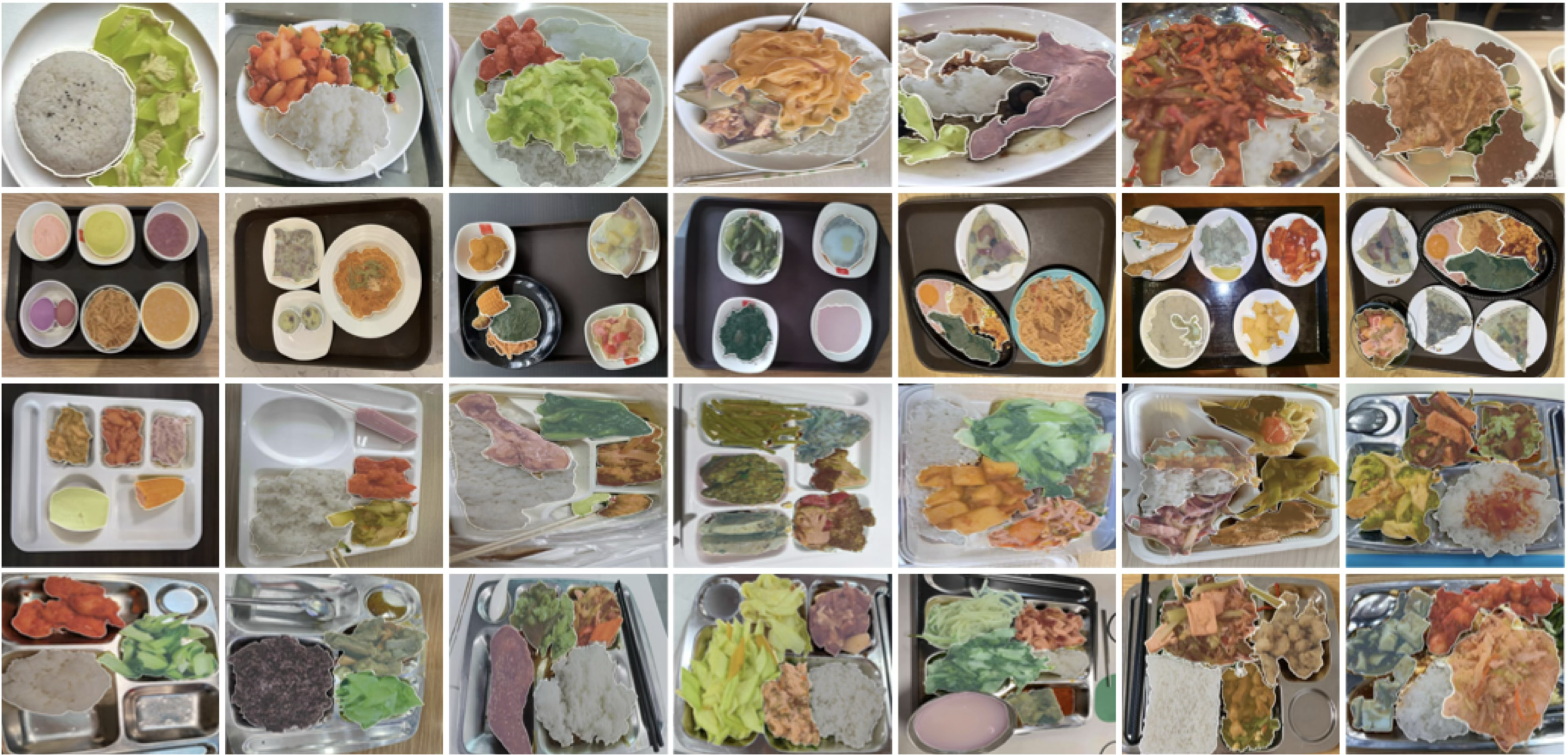}
    \caption{Representative annotated samples from DishSeg24k across diverse serving styles, which range from single plates to compartment trays and multi-section metal trays. Semi-transparent colored regions denote dish annotations.}
    \Description{DishSeg24k images from multiple dining settings with colored overlays marking individual dish regions.}
   \label{fig:our_dataset_vis}
\end{figure*}

Food segmentation, which assigns a category label to every food-relevant region in an image, is important for intelligent catering systems in restaurants and canteens. These systems price meals on a per-dish basis, requiring each dish in a dining image to be accurately recognized. Moreover, precise dish-level masks enable volume estimation of individual food items~\cite{almughrabi2025voltex}, which in turn supports dietary assessment and personalized nutritional recommendation~\cite{abhilash2025nutrivision}.
Unlike general-purpose object segmentation, food segmentation must resolve the compositional complexity of real-world dining scenes. Dishes are not isolated rigid objects, but visually entangled ensembles of mixed ingredients served in shared, cluttered spaces~\cite{min2023large,wu2021foodseg103}.
This compositional nature creates three challenges. 
(i)~\textit{Dense inter-dish overlap.} Multiple dishes are often served in shared containers, compartment trays, or closely packed plates. The resulting spatial overlap and adjacency make it difficult to delineate individual dish masks. For example, several stir-fried dishes may occupy the same tray section, creating heavily overlapping regions (e.g., row~3, col.~2 in Fig.~\ref{fig:our_dataset_vis}).
(ii)~\textit{Fine-grained class similarity.} Different dish classes often share common ingredients. For instance, both ``scrambled eggs with tomato'' and ``stir-fried beef with tomato'' contain tomato, producing locally indistinguishable visual features. Even when spatial boundaries are clear, the model must determine which class a region belongs to from limited local evidence. 
(iii)~\textit{Long-tail class distribution.} The natural consumption frequency of dishes follows a severe long-tail pattern. Common staples (e.g., \textit{Staple-Rice}) dominate the training set, while rare regional specialties may have few instances. This skewed distribution biases models toward frequent classes.

\begin{figure}[t]
    \centering
    \includegraphics[trim=5 5 5 5,clip,width=0.9\linewidth]{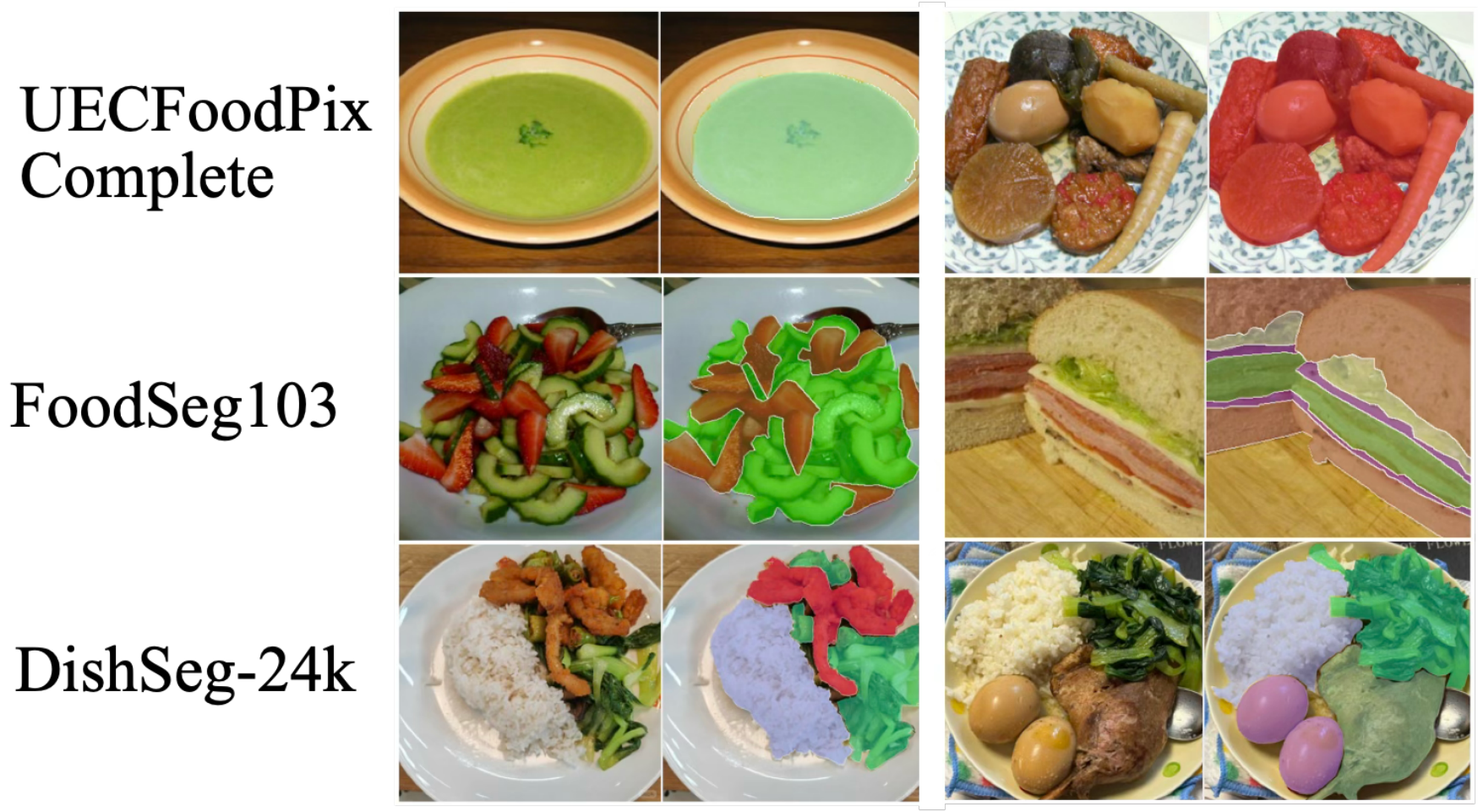}
    \caption{Cross-dataset qualitative comparison. For each dataset, we show images and corresponding segmentation masks.}
    \Description{Example food images and their segmentation masks from several food segmentation datasets, illustrating differences in scene complexity and annotation granularity.}
  \label{fig:dataset_vis_compare}
\end{figure}

Existing methods across multiple paradigms also fall short. CNN-based methods~\cite{chen2018deeplabv3plus} lack global context for dense overlap. Multi-modal pre-training~\cite{wu2021foodseg103}, self-supervised learning~\cite{liu2023feature}, and open-vocabulary approaches~\cite{wu2024ovfoodseg} improve representations yet remain too coarse for pixel-level boundary disambiguation. Foundation-model adaptations~\cite{lan2023foodsam} introduce segmentation capabilities but lack category-level discrimination under class imbalance.

Query-based frameworks such as Mask2Former~\cite{cheng2022mask2former} rely on deterministic decoding, which suppresses hypothesis diversity and limits robustness in ambiguous food scenes. This limitation is exacerbated by their use of a single static feed-forward network (FFN), whose fixed capacity cannot adequately model heterogeneous dish textures. A natural remedy is to introduce MoE routing, but under long-tail data distributions, naive MoE training often leads to expert collapse~\cite{chi2022moerepresentation,wu2024moemulti}.

Existing benchmarks are limited to clearly separated single-dish layouts~\cite{okamoto2021uecfoodpix} or ingredient-level annotations~\cite{wu2021foodseg103}, failing to capture the complexity of real-world dining scenes. To address this gap, we first introduce DishSeg24k, a large-scale dish-level segmentation benchmark constructed from real-world dining scenes and comprising 24,096 images, 112,281 instances, and 278 fine-grained categories. The dataset was created through multiple rounds of iterative annotation and rigorous multi-stage quality control, ensuring accurate segmentation boundaries and high annotation consistency across complex dining scenes. Fig.~\ref{fig:our_dataset_vis} illustrates the diversity of DishSeg24k. Each row corresponds to a distinct serving style (e.g., single plates, compartment trays, and multi-section metal trays), while the columns progress from simple to complex segmentation scenarios, revealing how the challenges intensify as scene composition becomes more complex. As shown in Fig.~\ref{fig:dataset_vis_compare}, DishSeg24k exhibits substantially denser inter-dish overlap and stronger fine-grained class similarity than UECFoodPixComplete~\cite{okamoto2021uecfoodpix} and FoodSeg103~\cite{wu2021foodseg103}. 

Furthermore, we propose Food Expert-Adaptive Segmentation Transformers (FEAST), an RL-guided decoder designed for dish-level segmentation in complex dining scenes. FEAST reformulates query-based mask decoding as a sequential decision process, in which the update at each decoder layer is treated as an action in an MDP\@. This formulation enables object queries to explore multiple plausible semantic hypotheses before convergence, overcoming the rigid deterministic behavior that often fails in scenes with inter-dish overlap. To address boundary ambiguity from mixed ingredients, we replace the standard FFN with a Mixture-of-Experts (MoE) decoder, in which query states are adaptively routed to specialized experts. To stabilize this routing under long-tail data, we further introduce a decoupled dual-critic optimization scheme that separately supervises task-oriented mask refinement and structure-aware expert routing, encouraging expert specialization while preventing expert collapse.

Extensive experiments on DishSeg24k demonstrate the state-of-the-art performance of FEAST, with improvements of 3.21\% mIoU, 3.68\% mDice, and 4.00\% mAcc over the strongest baseline. To further assess its effectiveness, we evaluate FEAST on FoodSeg103~\cite{wu2021foodseg103}. Under the same ResNet-50 setting, FEAST achieves a 1.47\% mIoU improvement over existing approaches.

Our contributions are summarized as follows. (i) We introduce DishSeg24k, a large-scale dish-level food segmentation dataset designed to capture dense overlap, fine-grained class similarity, and long-tail distributions in real-world dining scenes. (ii) We propose FEAST, which formulates query-based decoding as an RL-optimized MDP and introduces an MoE decoder with a decoupled dual-critic optimization scheme to improve both uncertainty-aware mask refinement and structure-aware expert routing. (iii) We conduct comprehensive experiments on DishSeg24k by evaluating FEAST against 13 representative segmentation methods and further validate its effectiveness on FoodSeg103.

\section{Related Work}
\textbf{Food Segmentation Datasets.} Food datasets have progressed from classification~\cite{bolanos2017ingredients,chen2021multitaskingredient,alla2025foodlens} and detection~\cite{lan2022deep,pandey2022object,lv2024differential} to pixel-level segmentation. Table~\ref{tab:exist_food_dataset} shows that early efforts such as MyFood~\cite{freitas2020myfood} and Mixed-Dishes~\cite{wang2019mixeddish} captured realistic scenes but were small-scale or lacked fine-grained masks. Among benchmarks, UECFoodPixComplete~\cite{okamoto2021uecfoodpix} provides dish-level annotations but features cleanly separated items without meaningful overlap. Conversely, FoodSeg103~\cite{wu2021foodseg103} and FoodSeg154~\cite{wu2021foodseg103} capture denser spatial interactions but operate strictly at the ingredient level, failing to reflect the integrated culinary identity of mixed recipes. Recent datasets that explicitly model occlusion, such as SibNet~\cite{nguyen2022sibnet}, remain limited in scale and taxonomic granularity.
In summary, no existing benchmark simultaneously provides large-scale dish-level annotations and captures the severe inter-dish adjacency, boundary ambiguity, and long-tail distributions inherent in authentic dining scenes. 

To fill this gap, we introduce DishSeg24k, which contains 24,096 images, 112,281 instances, and 278 dish-level categories, with an average density of 4.66 instances per image. It provides a comprehensive testbed that faithfully reflects the compositional complexity of real-world meals.

\begin{table}[t!]
\centering
\caption{Overview of food segmentation datasets.}
\label{tab:exist_food_dataset}
\resizebox{\linewidth}{!}{
\begin{tabular}{l|c|ccc|c}
\toprule
Dataset & Year & \#Imgs. & \#Cls. & \#Ins. & Level \\
\midrule
Food-201 \cite{meyers2015im2calories} & 2015 & 12,093 & 201 & 29,000 & Dish \\
MyFood \cite{freitas2020myfood} & 2020 & 1,250 & 9 & 1,250 & Dish  \\
UECFoodPixComplete \cite{okamoto2021uecfoodpix} & 2021 & 10,000 & 102 & 47,100 & Dish \\
FoodSeg103 \cite{wu2021foodseg103} & 2021 & 7,118 & 103 & 26,016 & Ingredient  \\
FoodSeg154 \cite{wu2021foodseg103} & 2021 & 9,490 & 154 & 59,773 & Ingredient  \\
SibNet \cite{nguyen2022sibnet} & 2022 & 12,557 & 138 & 37,671 & Dish \\
MyFoodRepo-273 \cite{mohanty2022myfoodrepo} & 2022 & 24,119 & 273 & 39,325 & Dish\\
MixedDishes \cite{nguyen2024foodmask_mixdishes2} & 2024 & 9,254 & 184 & 39,668 & Dish \\ \midrule
\textbf{DishSeg24k (Ours)} & 2026 & 24,096 & 278 & 112,281 & Dish \\
\bottomrule
\end{tabular}}
\end{table}

\noindent\textbf{Food Segmentation Methods.}
Food segmentation methods broadly comprise general-purpose architectures and food-specific designs.
Among general-purpose methods, CNN-based approaches such as DeepLabV3+~\cite{chen2018deeplabv3plus} remain common baselines but are limited by fixed receptive fields.
Per-pixel Transformers address this limitation by modeling global context. SegFormer~\cite{xie2021segformer} combines a hierarchical encoder with a lightweight MLP decoder, while subsequent methods improve embedding efficiency~\cite{yu2024embeddingfree}, contextual reasoning~\cite{ni2024contextguided}, and omni-scale feature aggregation through state-space models (SegMAN~\cite{fu2025segman}).
A more fundamental shift is the mask-classification paradigm, in which MaskFormer~\cite{cheng2021maskformer} reformulates segmentation as predicting class-labeled binary masks via learnable queries.
Mask2Former~\cite{cheng2022mask2former} adds masked cross-attention and multi-scale query refinement, supporting task-conditioned unification (OneFormer~\cite{jain2023oneformer}), state-space encoder integration (VMFormer~\cite{yan2024vmformer}), and encoder-derived query initialization (FeedFormer~\cite{shim2023feedformer}).
However, across these architectures, query updates follow deterministic trajectories toward a single semantic hypothesis. This deterministic refinement limits hypothesis diversity in compositionally entangled food scenes.

Food-specific methods follow a distinct progression. Early efforts focus on representation enhancement: ReLeM~\cite{wu2021foodseg103} aligns visual features with recipe-language embeddings to reduce intra-class variance, and FeaSC~\cite{liu2023feature} improves discriminability through self-supervised contrastive pre-training. Subsequent efforts shift toward transferability and category scalability: FoodSAM~\cite{lan2023foodsam} adapts SAM for zero-shot food segmentation, while OVFoodSeg~\cite{wu2024ovfoodseg} introduces open-vocabulary recognition via text--image alignment. More recently, FoodMask~\cite{nguyen2024foodmask_mixdishes2} moves toward unified multi-task modeling by jointly handling counting, segmentation, and recognition. Although these advances strengthen food-domain representations and category coverage, they still rely on deterministic query refinement and do not explicitly resolve the visual-semantic ambiguity caused by inter-dish overlap and inter-class similarity. FEAST addresses this limitation by casting query refinement as an RL-guided stochastic decision process with MoE routing for active hypothesis exploration.

\begin{figure*}[t!]
    \centering
    \begin{subfigure}[t]{0.6\linewidth}
        \centering
        \includegraphics[trim=5 5 5 5,clip,width=\linewidth]{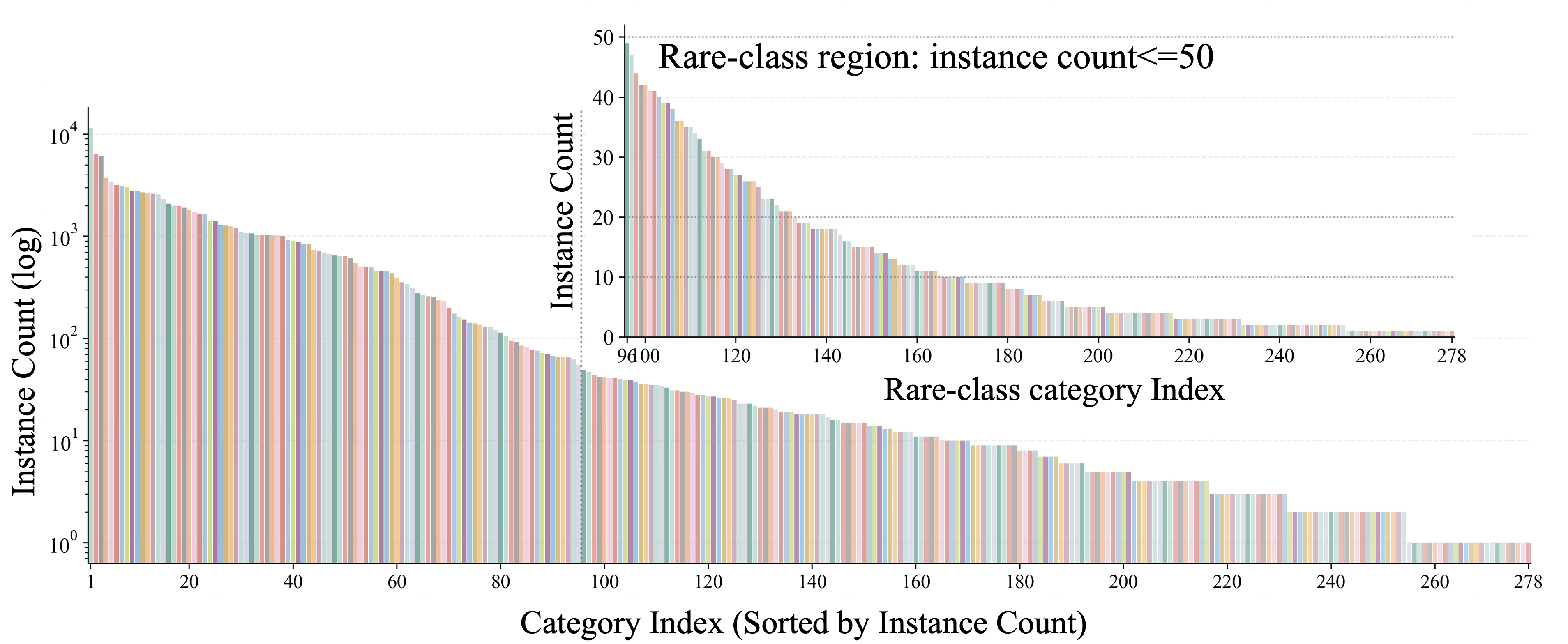}
        \caption{}
        \label{fig:long_tail}
    \end{subfigure}
    \hfill
    \begin{subfigure}[t]{0.35\linewidth}
        \centering
        \includegraphics[trim=5 50 5 50, clip, width=\linewidth]{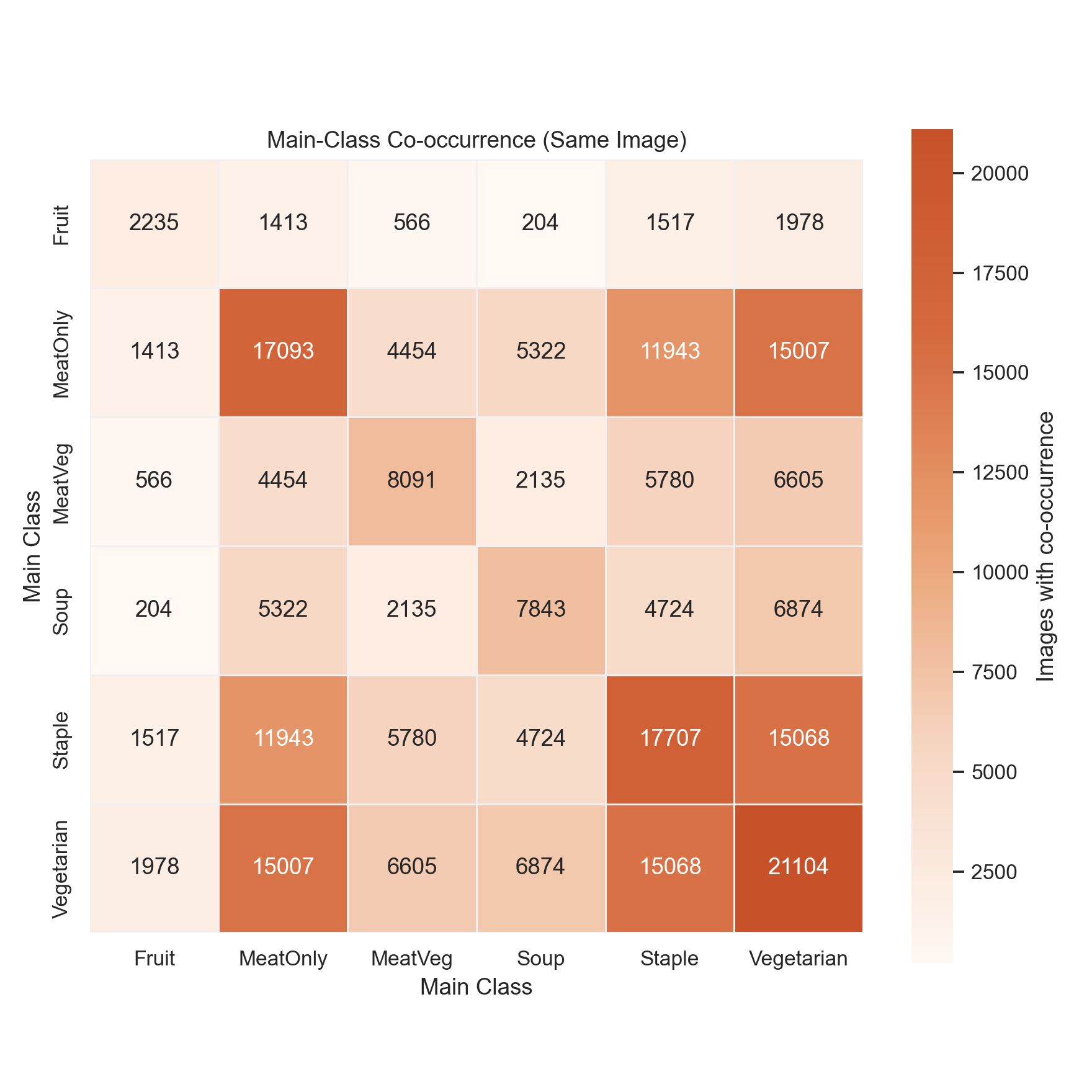}
        \caption{}
        \label{fig:cooccurrence}
    \end{subfigure}
    \caption{Statistical analysis of DishSeg24k. (a) The category distribution exhibits a pronounced long-tail pattern. The upper-right panel provides a linear-scale view of rare classes, while the lower panel shows the full distribution on a logarithmic scale. (b) Co-occurrence statistics reveal high scene complexity with frequent interactions among multiple categories.}
    \Description{A long-tail category-frequency chart and a category co-occurrence heatmap summarizing the class imbalance and dense interactions in DishSeg24k.}
    \label{fig:dataset_stats}
\end{figure*}

\section{DishSeg24k Dataset}\label{sec:dishseg24k_dataset}
Below, we describe the collection process, annotation protocol, and key statistical properties that distinguish DishSeg24k from existing benchmarks.

\subsection{Dataset Collection}
We build our dataset from two complementary sources. First, we collect 3,493 real-world dining images from publicly available online sources. This subset focuses on Chinese dining settings, where dishes are more often shared and visually entangled, resulting in complex mixtures, heavy occlusions, and ambiguous boundaries. Second, we adopt images from the ZSFood dataset~\cite{zhou2023ZSFood}, which contains 20,603 food images across 10 restaurant scenarios, with 95,322 bounding boxes covering 291 classes. Notably, ZSFood includes both Western-style and Chinese-style dishes, providing diverse yet relatively structured dining scenes. We discard the original bounding-box annotations and perform complete pixel-level re-annotation using polygon masks under our unified hierarchical labeling protocol. All images undergo quality filtering to ensure sufficient visual clarity and contextual realism. The resulting dataset spans restaurants, canteens, and home environments, featuring varied dish arrangements ranging from isolated plates to highly cluttered scenes with shared containers. By combining structured scenes from ZSFood with highly complex Chinese dining cases, our dataset provides a more comprehensive and challenging benchmark for dish-level food segmentation.

\subsection{Annotation Protocol}
\textbf{Label Taxonomy Design.}\footnote{Please refer to the supplementary material for the full taxonomy and category list.}
We design a hierarchical taxonomy for food segmentation using a composition-based naming convention: [Category]--[Ingredient]--[Optional Cooking Method] (e.g., Meat-Veg--Beef--Potato). Dishes are organized into six major categories---Staple, Soup, Fruit, Vegetarian, Meat-Only, and Meat-Veg---and classified into 278 dish-level labels defined by their key ingredients. Cooking methods are included only when different preparations of the same ingredients produce visually distinct appearances (e.g., steamed versus braised), with stir-frying treated as the default. Although our dataset focuses on dish-level segmentation, the additional ingredient-level annotations enable broader downstream applications such as nutritional analysis and dietary assessment.

\noindent\textbf{Annotation Workflow.}
We employ a multi-stage annotation process on the X-AnyLabeling~\cite{X-AnyLabeling} platform. First, six expert annotators spent two months in a pilot phase, labeling 2,000 samples to establish boundary criteria and refine taxonomy definitions. The production phase then involved 15 annotators over two additional months. To balance efficiency and precision, we use AI-assisted labeling. Specifically, annotators refine initial masks generated by foundation models (e.g., SAM~\cite{kirillov2023sam}) to achieve pixel-level accuracy. For cluttered scenes, each visually distinguishable dish receives an individual mask, while overlapping ingredients are grouped into a unified mask based on the dominant dish identity to maintain semantic coherence. Finally, a three-tier verification process ensures annotation quality: (i) annotators self-check their annotations against the guidelines; (ii) the quality assurance team samples 20\% of the weekly output and rejects batches with pixel-level accuracy below 95\% for re-annotation; and (iii) domain experts conduct final reviews to correct category misclassifications and boundary inconsistencies.

\subsection{Statistics and Analysis}\par
\textbf{Basic Statistics.}
DishSeg24k comprises 24,096 images with 112,281 pixel-level instance annotations across 278 fine-grained categories and six super-categories: Staple, Soup, Fruit, Vegetarian, Meat-Only, and Meat-Veg. Unlike existing benchmarks, DishSeg24k captures the inherent clutter of real-world dining, exhibiting a high instance density of 4.66 instances per image and an average of 4.43 unique categories per image.

\noindent\textbf{Distribution Analysis and Composition Complexity.} Fig.~\ref{fig:dataset_stats} summarizes the statistical properties of DishSeg24k in terms of category distribution and meal composition. Fig.~\ref{fig:dataset_stats}(a) shows that the dataset exhibits a long-tail distribution, in which a small fraction of categories dominate while many others are underrepresented. This imbalance reflects realistic dining patterns and poses significant challenges for representation learning under class scarcity. Fig.~\ref{fig:dataset_stats}(b) further shows that frequent co-occurrence patterns (e.g., Staple--Vegetarian and Meat-Only--Staple) confirm the compositional nature of real-world meals.

\noindent\textbf{Comparison with Existing Food Segmentation Datasets.} 
Fig.~\ref{fig:dataset_vis_compare} compares DishSeg24k with representative food segmentation datasets. While UECFoodPixComplete~\cite{okamoto2021uecfoodpix} offers dish-level masks, its scale and scene diversity remain limited. FoodSeg103~\cite{wu2021foodseg103} provides ingredient-level annotations but does not capture dish-level structures. As summarized in Table~\ref{tab:exist_food_dataset}, DishSeg24k is substantially larger across the key dataset dimensions. 

\begin{figure*}[t!]
  \centering 
  \includegraphics[width=0.8\textwidth]{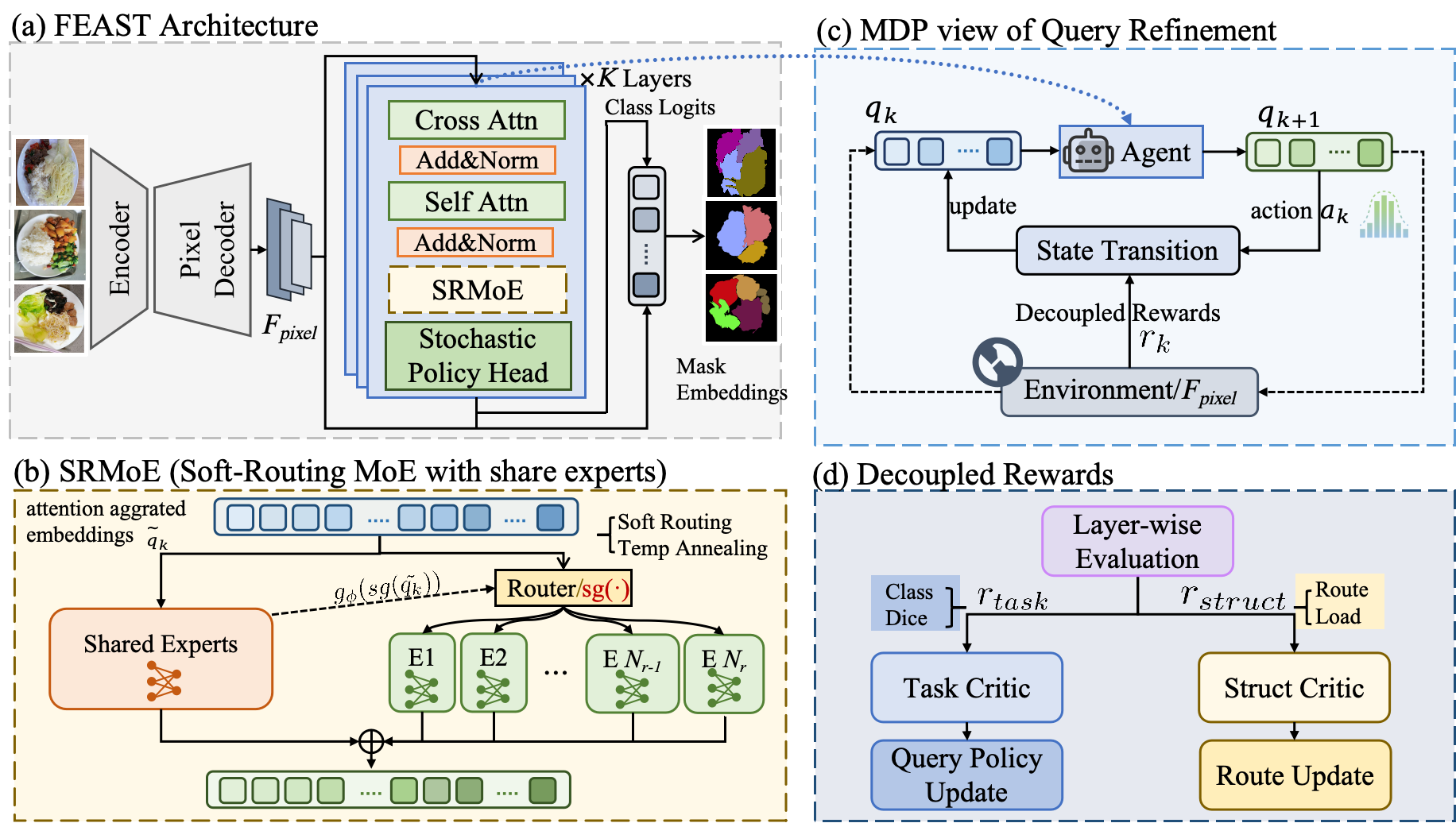}
  \caption{Overview of FEAST. (a) Overall pipeline. (b) Stochastic MoE module within a decoder layer. (c) MDP for query refinement. (d) Decoupled rewards in the MDP.}
  \Description{Architecture diagram of FEAST showing the segmentation pipeline, stochastic mixture-of-experts decoder, Markov decision process, and dual reward signals.}
  \label{fig:feast_framework}
\end{figure*}

\section{FEAST Method}
\subsection{Overview}\label{sec:overview}
Fig.~\ref{fig:feast_framework}(a) shows the overall architecture of FEAST, which consists of three primary components: a visual feature encoder, a pixel decoder, and a novel stochastic Transformer decoder. Given an input dining image $I \in \mathbb{R}^{3 \times H \times W}$, the visual feature encoder first extracts multi-scale pixel embeddings, which are subsequently fed into the pixel decoder to produce multi-scale mask features $F_{\text{pixel}}$. A set of $N$ learnable object queries is then iteratively refined through $K$ Transformer decoder layers. Following the standard Mask2Former pipeline~\cite{cheng2022mask2former}, prediction heads project the refined queries into class logits and binary masks.
Our key departure is to reformulate the deterministic query update across decoder layers as a \emph{stochastic sequential decision process}. 
At each decoder layer $k$, a \textbf{MoE} module replaces the standard FFN to produce specialized features (Fig.~\ref{fig:feast_framework}(b)), and a \textbf{stochastic policy} samples a continuous semantic shift as the query update (Fig.~\ref{fig:feast_framework}(c)).
A \textbf{decoupled dual-critic} optimization scheme provides separate reward signals to guide mask refinement and expert routing (Fig.~\ref{fig:feast_framework}(d)).

\subsection{MDP Formulation for Query Refinement}\label{sec:mdp}
We model the progressive update of object queries across $K$ decoder layers as a finite-horizon MDP\@. A single forward pass for one image constitutes a complete episode of $K$ steps. We define the MDP as a tuple $\mathcal{M}{=}(\mathcal{S},\mathcal{A},\mathcal{P},\mathcal{R},\gamma)$.

\textbf{State Space} $\mathcal{S}$.
The state at step $k$ is the set of object queries $s_k = \{q_k^1, q_k^2, \ldots, q_k^N\} \in \mathbb{R}^{N\times D}$. The pixel features $F_{\text{pixel}}$ serve as a \emph{static context} rather than a state variable. Under this contextual MDP formulation~\cite{hallak2015contextual}, the Markov property holds conditioned on the fixed context.

\textbf{Action Space} $\mathcal{A}$.
An action $a_k \in \mathbb{R}^{N\times D}$ represents a continuous semantic shift in the query embedding space. Each of the $N$ queries is updated independently, thereby decomposing the joint action into $N$ parallel sub-problems of dimension $D$.

\textbf{Transition Function} $\mathcal{P}$.
The transition is deterministic and follows a residual update rule: $q_{k+1} = q_k + a_k$.
This residual formulation admits a clear semantic interpretation: the action $a_k$ encodes a \emph{semantic correction} that shifts the query representation toward a refined segmentation hypothesis. When $a_k{\approx}0$, the query preserves its current semantic content, while larger actions enable exploratory jumps to alternative hypotheses.

\textbf{Reward Function} $\mathcal{R}$.
The stepwise reward $r_k$ measures the incremental improvement in segmentation quality from layer $k{-}1$ to $k$. We design a dual-reward structure comprising a \emph{task reward} that evaluates improvements in mask quality and a \emph{structure reward} that assesses routing efficiency. This design enables decoupled optimization of the query policy and router.
 
\textbf{Discount Factor} $\gamma$.
We use $\gamma{=}1$ (undiscounted), as the short horizon ($K{=}9$ layers) renders discounting unnecessary.
 
The objective is to learn a stochastic policy $\pi_\theta(a_k\!\mid\!s_k)$ that maximizes the expected cumulative reward $\mathbb{E}_{\pi_\theta}[\sum_{k=1}^{K} r_k]$, enabling queries to explore multiple semantic hypotheses rather than converging along a single deterministic trajectory.

\subsection{Stochastic MoE Policy Network}\label{sec:moe}
\textbf{Feature extraction via MoE.}
At layer $k$, queries first attend to image features through masked cross-attention and then aggregate inter-query context via self-attention, producing an attention-aggregated feature $\tilde{q}_k \in \mathbb{R}^{N\times D}$. We replace the standard FFN with an MoE module consisting of one \emph{shared expert} $f_{\text{s}}$ that is always active and $E_r$ \emph{routed experts} $\{f_e\}_{e=1}^{E_r}$. The shared expert captures task-agnostic visual primitives (e.g., edge and texture statistics common across all dish categories), while the routed experts specialize in category-specific patterns (e.g., liquid surfaces vs.\ solid textures).

\textbf{Decoupled soft routing.}\label{sec:routing}
The query embeddings serve dual roles as semantic memory and routing input.
To prevent the routing gradient from biasing queries toward router-specific rather than task-relevant features, we apply a \textbf{stop-gradient} operator $\mathrm{sg}(\cdot)$ on the routing input.
The routing probability for expert $e$ is:
\begin{equation}\label{eq:routing}
  g_\phi(e \mid \tilde{q}_k) = \mathrm{softmax}\!\Big(\frac{W_r\,\mathrm{sg}(\tilde{q}_k)}{\tau}\Big),
\end{equation}
where $W_r$ denotes the router parameters and $\tau$ is a temperature that is linearly annealed from $1.0$ to $0.01$ during training.
At inference time, the annealed routing naturally reduces to Top-1 hard selection.
The MoE output is a weighted combination of the expert outputs:
\begin{equation}\label{eq:moe_output}
  h_k = f_{\text{s}}(\tilde{q}_k) + \sum_{e=1}^{E_r} 
g_\phi(e \mid \tilde{q}_k)\; f_e(\tilde{q}_k).
\end{equation}

\textbf{Stochastic action sampling.}
Given the MoE feature $h_k$, the policy network outputs a Gaussian distribution over actions:
\begin{equation}\label{eq:action}
  \mu_k = W_\mu\, h_k,\quad
  \sigma_k = \mathrm{softplus}(W_\sigma\, h_k) + \epsilon,\quad
  a_k \sim \mathcal{N}\!\big(\mu_k,\;\mathrm{diag}(\sigma_k^2)\big),
\end{equation}
where $\epsilon{=}10^{-6}$ ensures numerical stability. The state is then updated as $q_{k+1} = q_k + a_k$. At inference, we set $a_k{=}\mu_k$.

\subsection{Reward Decoupling and Dual-Critic Optimization}\label{sec:reward_decoupling}
Optimizing the stochastic MoE decoder presents a challenge in credit assignment: when segmentation quality improves, the improvement may be attributed to a better query shift (policy) or a better expert selection (router).
Moreover, the stop-gradient operator decouples routing from query-feature optimization, motivating a dedicated structure-aware RL objective for the routing parameters $\phi$ rather than relying solely on segmentation backpropagation.
We address both issues through decoupled reward signals and dual-critic optimization.

\textbf{Target assignment.}
Following Mask2Former, we compute the optimal bipartite matching $\hat{\sigma}$ between queries and ground-truth objects at the final layer $K$, and apply it consistently across all layers to obtain per-query, per-step evaluation targets.

\textbf{Task reward.}
For each matched query $j$ ($j{=}1,\dots,M$, where $M$ is the number of matched pairs), the task reward measures the stepwise improvement in mask quality and classification confidence from layer $k{-}1$ to $k$:
\begin{equation}\label{eq:task_reward}
  r_k^{j,\mathrm{task}} \;=\; w_{c_j}\!\Big[
    \lambda_{\mathrm{dice}}\;\Delta\mathrm{Dice}_k^{\,j}
    \;+\; \lambda_{\mathrm{cls}}\;\Delta\mathrm{Conf}_k^{\,j}
  \Big],
\end{equation}
where $\Delta\mathrm{Dice}_k^{\,j} = \mathrm{Dice}(M_k^{\hat{\sigma}(j)}, M_j^{\mathrm{gt}}) - \mathrm{Dice}(M_{k-1}^{\hat{\sigma}(j)}, M_j^{\mathrm{gt}})$ measures the improvement in mask overlap, $\Delta\mathrm{Conf}_k^{\,j} = p_k^{\hat{\sigma}(j)}(c_j) - p_{k-1}^{\hat{\sigma}(j)}(c_j)$ measures the improvement in classification confidence, and $w_{c_j}{=}1/\!\sqrt{n_{c_j}}$ reweights the reward by the inverse square root of the class frequency $n_{c_j}$ to address the long-tail distribution. Unmatched queries receive zero reward.

\textbf{Structure reward.}
The structure reward evaluates the router's load-balancing efficiency, independent of mask quality:
\begin{equation}\label{eq:struct_reward}
  r_k^{\mathrm{struct}} \;=\; \lambda_{\mathrm{div}}\, H(g_\phi)
  \;-\; \lambda_{\mathrm{bal}}\, \mathrm{KL}(\bar{p}_e \| U),
\end{equation}
where $H(g_\phi)$ is the per-query routing entropy, which encourages diverse expert usage, and $\mathrm{KL}(\bar{p}_e\|U)$ penalizes deviations of the batch-level expert activation frequency $\bar{p}_e$ from a uniform distribution $U$.
This reward acts as a tiebreaker: among routing strategies yielding comparable task performance, it favors the most balanced one, preventing expert collapse without interfering with segmentation quality.

\textbf{Dual-critic advantage estimation.}
Two independent critics estimate value functions for the decoupled rewards. (i) $V_{\mathrm{task}}(h_k)$ predicts the expected cumulative task reward from the feature $h_k$. (ii) $V_{\mathrm{struct}}(\mathrm{sg}(h_k))$ predicts the expected cumulative structure reward, while the stop-gradient ensures that this critic does not influence the query features.
Advantages $A_k^{\mathrm{task}}$ and $A_k^{\mathrm{struct}}$ are computed using generalized advantage estimation (GAE)~\cite{schulman2015gae} with $\lambda_{\mathrm{GAE}}{=}0.95$.

\textbf{Policy optimization.}
Because each forward pass constitutes a complete $K$-step episode and the parameters are updated once per batch, data collection and optimization share the same policy parameters. Under this on-policy setting, the query policy $\theta$ is optimized via a policy-gradient objective with a KL regularization term that anchors it to the reference policy $\pi_{\mathrm{ref}}$ derived from the pretrained Mask2Former decoder:
\begin{equation}\label{eq:pg}
  J_{\mathrm{query}}(\theta)
    = \mathbb{E}\!\Big[
        \log \pi_\theta(a_k \!\mid\! s_k)\; A_k^{\mathrm{task}}
      \Big]
    - \beta\,\mathrm{KL}\!\big(\pi_\theta \,\|\, \pi_{\mathrm{ref}}\big),
\end{equation}
where $A_k^{\mathrm{task}}$ is the task advantage estimated by $V_{\mathrm{task}}$, $\beta{=}0.01$ is the KL penalty coefficient, and $\pi_{\mathrm{ref}}$ is the reference policy derived from the pretrained Mask2Former decoder. The reference policy remains frozen throughout training to prevent catastrophic deviation from the deterministic baseline. The KL penalty regularizes policy updates analogously to the penalty formulation of constrained optimization~\cite{schulman2017ppo}, bounding deviation from the reference without an explicit clipping ratio.

The routing policy $\phi$ is optimized via REINFORCE driven by the structure advantage:
\begin{equation}\label{eq:route}
  J_{\mathrm{route}}(\phi) = \mathbb{E}\!\Big[
    \log g_\phi\!\big(e \mid \mathrm{sg}(\tilde{q}_k)\big)\; A_k^{\mathrm{struct}}
  \Big].
\end{equation}
Because $\mathrm{sg}(\cdot)$ prevents gradients from flowing into the query features, $\nabla_\phi J_{\mathrm{route}}$ provides an unbiased policy gradient solely for the router parameters.

\subsection{Overall Training Objective}\label{sec:objective}
The total loss combines a segmentation loss and two RL objectives:
\begin{equation}\label{eq:total}
  \mathcal{L}_{\mathrm{total}}
    = \mathcal{L}_{\mathrm{seg}}
    - \big(J_{\mathrm{query}} + J_{\mathrm{route}}\big).
\end{equation}
Here, $\mathcal{L}_{\mathrm{seg}}$ is the standard Mask2Former loss (per-query cross-entropy, Dice, and binary cross-entropy) evaluated at the final decoder layer $K$. The optimal bipartite matching $\hat{\sigma}$ is computed once at layer $K$ and reused across all layers to ensure consistent supervision.
The three objective components are complementary: (i) $\mathcal{L}_{\mathrm{seg}}$ provides dense per-pixel gradients for stable convergence; (ii) $J_{\mathrm{query}}$ guides global trajectory optimization for mask quality; and (iii) $J_{\mathrm{route}}$ maintains routing diversity to prevent expert collapse.
All parameters are optimized jointly using a single AdamW optimizer. Since each forward pass constitutes one complete $K$-step episode, no separate RL update loop is required; the RL objectives integrate seamlessly into standard end-to-end training.

\section{Experiments}\par
\subsection{Experimental Settings}
\noindent\textbf{Datasets.}
We evaluate FEAST on two benchmarks.
(i)~DishSeg24k comprises 24,096 images (19,365/4,731 train/test) with 112,281 pixel-level annotations spanning 278 dish categories in real-world dining scenes. It features a long-tail distribution, high instance density, and severe inter-dish overlap.
(ii)~We further evaluate FEAST on FoodSeg103~\cite{wu2021foodseg103} to assess the generalizability of our method.

\noindent\textbf{Evaluation Metrics.} We report mean Intersection-over-Union (mIoU), mean Dice coefficient (mDice), and mean class accuracy (mAcc) on the respective test sets.
 
\noindent\textbf{Implementation Details.}
FEAST uses Detectron2 with ResNet-50 and Swin-B backbones pretrained on ImageNet~\cite{wu2019detectron2,he2016resnet,liu2021swin,deng2009imagenet}. We use AdamW with a learning rate of $1{\times}10^{-4}$, a weight decay of $0.05$, and a step-decay schedule. Models are trained for 80,000 iterations using $512{\times}512$ crops and a batch size of 16 on eight NVIDIA RTX 3090 GPUs and are tested at the original resolution. The decoder contains $K{=}9$ layers, $N{=}100$ queries, and $E_r{=}4$ experts with Top-2 soft routing. The frozen, pretrained Mask2Former decoder serves as $\pi_{\mathrm{ref}}$, and advantages are estimated using GAE~\cite{schulman2015gae}. Baselines use official code and the same training protocol for each backbone.

\subsection{Quantitative Analysis}
To evaluate segmentation accuracy and computational efficiency on DishSeg24k, Table~\ref{tab:dishseg_all} compares FEAST with 13 representative methods. Among the baselines, query-based approaches consistently outperform CNN-based and per-pixel Transformer methods; for example, Mask2Former achieves 50.22\% mIoU, compared with 44.81\% for ED-AFormer. Food-specific methods remain less effective despite domain adaptation or larger backbones, suggesting that stronger representations alone do not resolve severe spatial and semantic ambiguity.

FEAST achieves the best results with both backbones. With ResNet-50, it reaches 53.43\% mIoU and surpasses Mask2Former by 3.21\% mIoU, 3.68\% mDice, and 4.00\% mAcc, while requiring 65.04M parameters and 75.94 GFLOPs. With Swin-B, FEAST obtains 55.07\% mIoU, 74.26\% mDice, and 63.36\% mAcc. These results demonstrate the effectiveness of RL-guided MoE decoding for compositional food scenes.
\begin{table}[t!]
\centering
\caption{Comparison with representative segmentation methods on \textbf{DishSeg24k} (\%).}
\label{tab:dishseg_all}
\resizebox{\linewidth}{!}{
\begin{tabular}{l|c|cc|ccc}
\toprule
Method & Backbone & Params (M) & GFLOPs 
       & mIoU & mDice & mAcc \\
\midrule
\multicolumn{7}{l}{\textit{CNN-based}} \\
DeepLabV3+~\cite{chen2018deeplabv3plus} 
& ResNet-50 & 26.75 & 77.72 & 32.89 & 37.76 & 57.64 \\

\midrule
\multicolumn{7}{l}{\textit{Per-pixel Transformer}} \\
SegFormer~\cite{xie2021segformer} 
& ResNet-50 & 24.88 & 29.69  & 38.96  & 43.63 & 47.10 \\
ED-AFormer~\cite{yu2024embeddingfree} 
& ResNet-50 & 113.98 & 59.31 & 44.81 & 50.59 & 56.52 \\
CGRSeg~\cite{ni2024contextguided} 
& ResNet-50 & 200.76 & 46.17 & 36.13 & 42.13   & 49.46 \\
SegMAN~\cite{fu2025segman} 
& ResNet-50 & 51.82 & 133.40 & 39.13 & 44.52 & 62.52 \\

\midrule
\multicolumn{7}{l}{\textit{Query-based Mask Classification}} \\
MaskFormer~\cite{cheng2021maskformer} 
& ResNet-50 & 41.29 & 58.53 & 50.05 & 56.98 & 57.32 \\
Mask2Former~\cite{cheng2022mask2former} 
& ResNet-50 & 44.02 & 93.17 & 50.22 & 57.30 & 56.92 \\
OneFormer~\cite{jain2023oneformer} 
& ResNet-50 & 64.86 & 89.26  & 49.24 & 57.25 & 57.13 \\
VMFormer~\cite{yan2024vmformer} 
& ResNet-50 & 49.89 & 91.13 & 48.31 & 55.51 & 55.25 \\
FeedFormer~\cite{shim2023feedformer} 
& ResNet-50 & 81.92 & 74.68 & 47.36 & 54.79 & 56.30 \\

\midrule
\multicolumn{7}{l}{\textit{Food-specific Methods}} \\
CCNet~\cite{wu2021foodseg103}
& ReLeM-ResNet50 & 615.28 & 71.36 & 34.20 & 46.97 & 59.74 \\
FoodSAM~\cite{lan2023foodsam}
& ViT-H & 460.13 & 632.75 & 37.90 & 43.64 & 50.38 \\
FDSNet~\cite{Xiao2025fdsnet}
& Swin-B & 102.30 & 550.62 & 30.27 & 34.73 & 35.38 \\

\midrule
\textbf{FEAST (Ours)}  
& ResNet-50 & 65.04 & 75.94 
& \textbf{53.43} & \textbf{60.98} & \textbf{60.92} \\
\textbf{FEAST (Ours)}  
& Swin-B & 127.98 & 188.40 & \textbf{55.07} & \textbf{74.26} & \textbf{63.36} \\

\bottomrule
\end{tabular}}
\end{table}

\subsection{Qualitative Analysis}
\begin{figure*}[t!]
  \centering
  \begin{subfigure}[t]{0.68\linewidth}
    \centering
    \includegraphics[trim=5 165 5 5,clip,width=\linewidth]{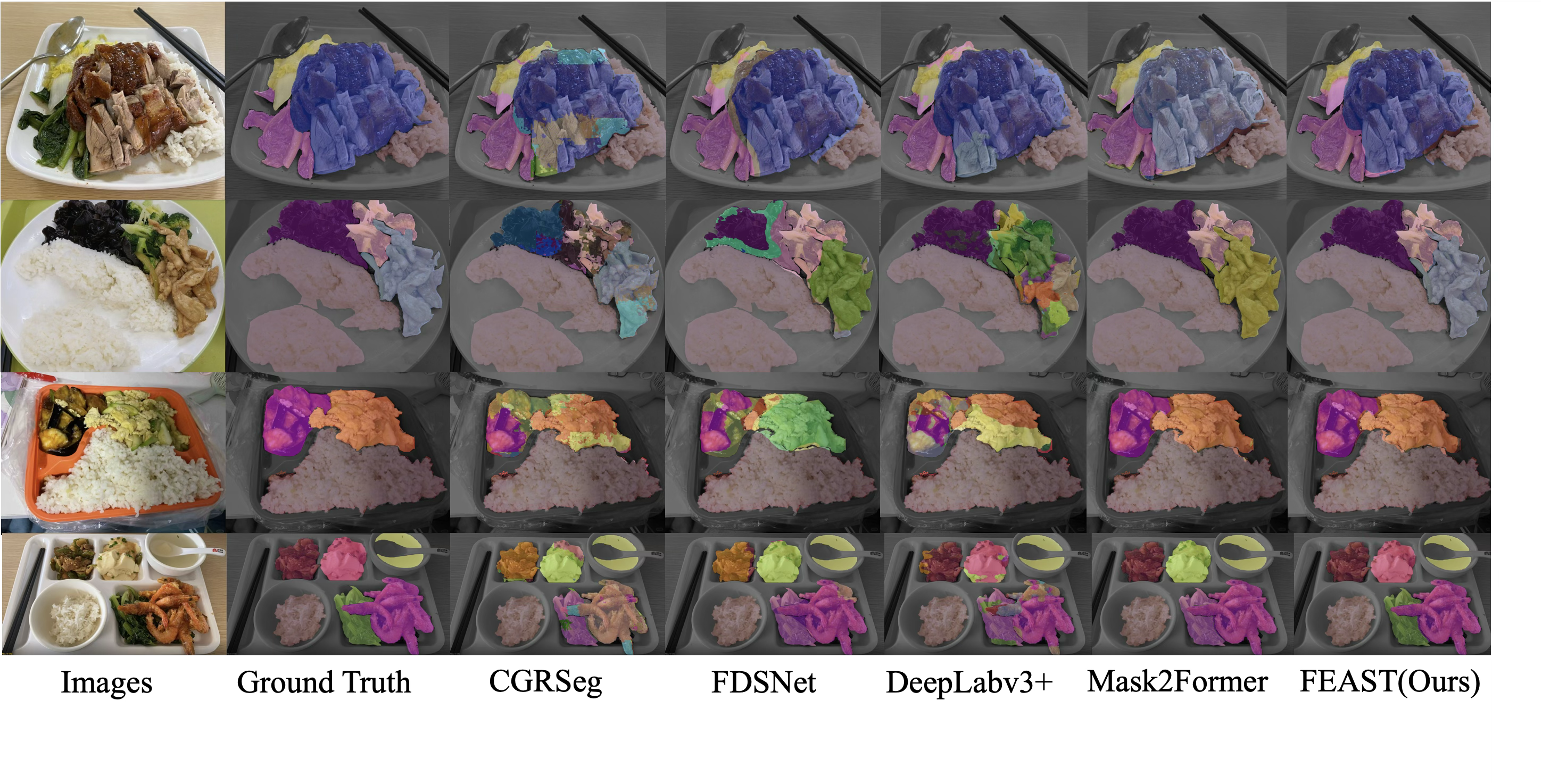}
    \caption{}
    \label{fig:seg_visual_sub}
  \end{subfigure}
  \hfill
  \begin{subfigure}[t]{0.3\linewidth}
    \centering
    \includegraphics[trim=5 20 5 5,clip,width=\linewidth]{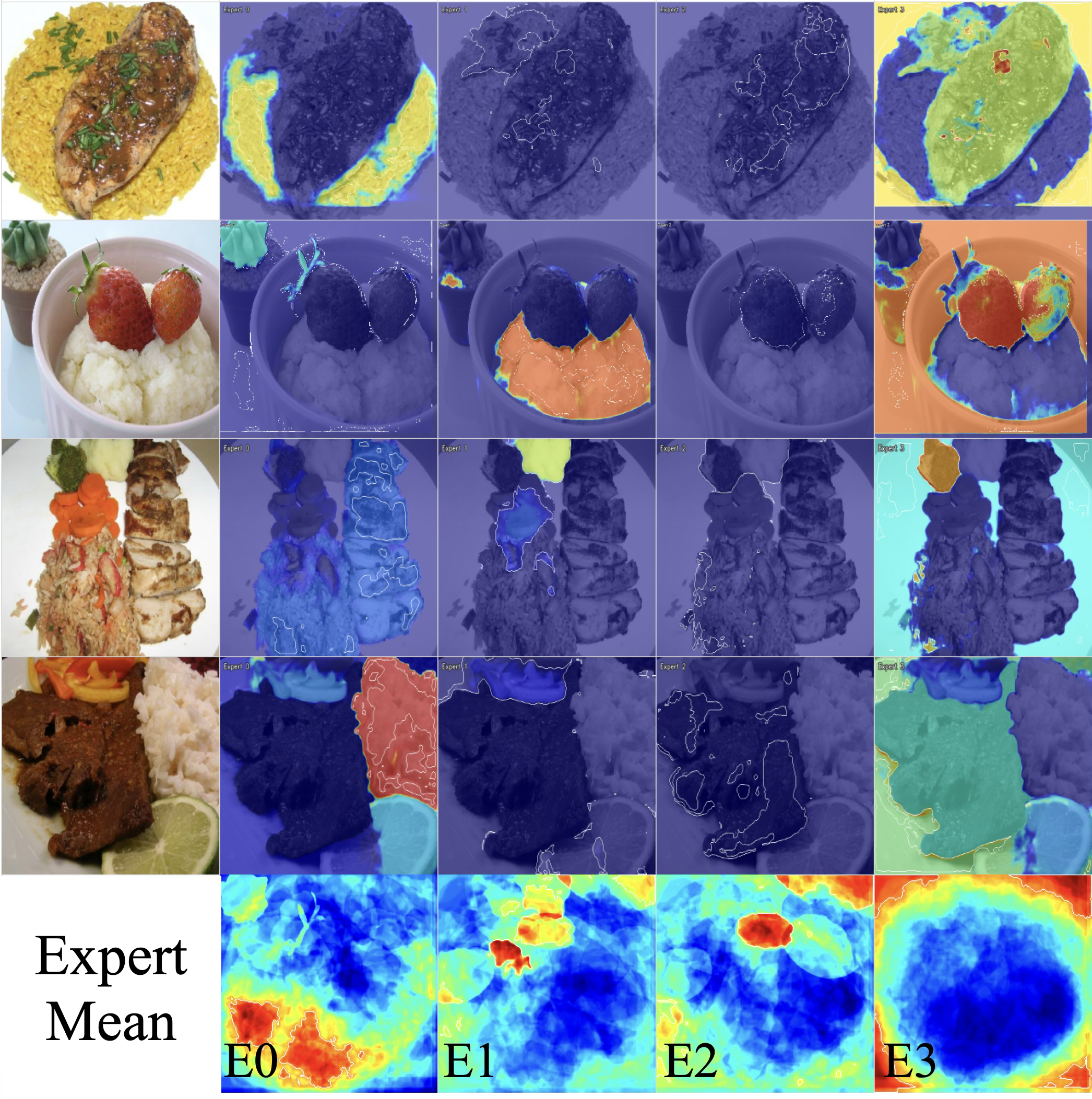}
    \caption{}
    \label{fig:moe_expert_sub}
  \end{subfigure}
  \caption{(a) Qualitative segmentation comparison between methods. (b) Activation maps of MoE experts.}
  \Description{Qualitative masks produced by several segmentation methods alongside heatmaps showing the spatial activation patterns of four FEAST experts.}
  \label{fig:seg_and_moe_combined}
\end{figure*}

To examine performance in ambiguous scenes and interpret expert routing, Fig.~\ref{fig:seg_and_moe_combined} presents segmentation results and expert activation maps. In Fig.~\ref{fig:seg_and_moe_combined}(a), the baselines suffer from fragmented regions, inaccurate boundaries, or category confusion, whereas FEAST produces clearer boundaries and more accurate labels. Fig.~\ref{fig:seg_and_moe_combined}(b) shows that Expert~0 focuses on the target dish, Expert~3 captures background context, and Experts~1--2 refine local boundaries. These stable patterns explain the qualitative improvements and indicate complementary expert specialization without explicit supervision.

\subsection{Ablation Study}
\textbf{Progressive Module Contribution.}
To isolate the cumulative contribution of each FEAST component, Table~\ref{tab:ablation_progressive} progressively adds the MoE decoder, query stochasticity, and RL guidance. The MoE decoder provides the largest individual gain (+3.03\% mIoU). Query stochasticity yields only a marginal additional improvement and slightly reduces mDice and mAcc, whereas RL guidance raises all metrics to their best values and improves mIoU by 3.21\% over the baseline. Thus, RL converts stochastic perturbation into controlled exploration.
\begin{table}[t!]
\centering
\caption{Progressive module contribution (\%).}
\label{tab:ablation_progressive}
\resizebox{\linewidth}{!}{
\begin{tabular}{c|l|ccc}
\toprule
ID & Configuration & mIoU & mDice & mAcc  \\
\midrule
A & Baseline                    & 50.22 & 57.30 & 56.92 \\
B & + MoE Decoder               & 53.25 & 60.48 & 60.68 \\
C & + MoE + Q-Stoch             & 53.28 & 59.99 & 60.33  \\
D & + MoE + Q-Stoch + Q-RL (FEAST) & \textbf{53.43} & \textbf{60.98} & \textbf{60.92} \\
\bottomrule
\end{tabular}}
\end{table}

\noindent\textbf{RL Training Strategy.}
To examine how the RL components interact, Table~\ref{tab:ablation_rl} evaluates query stochasticity, RL routing, and the dual critic with the MoE decoder fixed. Individual or pairwise configurations underperform the MoE-only reference (R0), indicating that partial RL optimization is unstable. Their joint configuration (R5) instead improves R0 by 0.18\% mIoU, 0.50\% mDice, and 0.24\% mAcc. The components are therefore complementary and must be jointly optimized to produce consistent gains.
\begin{table}[t!]
  \caption{Ablation on RL components (\%) with the MoE decoder fixed and enabled. R0 serves as the MoE-only reference.}
  \label{tab:ablation_rl}
  \centering
  \small
  \begin{tabular}{c|ccc|ccc}
    \toprule
    ID & Q-Stoch & RL Route & Dual Critic & mIoU & mDice & mAcc  \\
    \midrule
    R0 & $\times$      & $\times$      & $\times$      & 53.25 & 60.48 & 60.68 \\
    R1 & $\times$      & $\times$      & $\checkmark$  & 52.40 & 60.13 & 59.53 \\
    R2 & $\times$      & $\checkmark$  & $\times$      & 51.95 & 58.71 & 59.04 \\
    R3 & $\checkmark$  & $\times$      & $\checkmark$  & 52.07 & 58.91 & 59.12 \\
    R4 & $\times$      & $\checkmark$  & $\checkmark$  & 51.84 & 59.03 & 59.85\\ 
    \midrule
    R5 & $\checkmark$  & $\checkmark$  & $\checkmark$  & \textbf{53.43} & \textbf{60.98} &\textbf{60.92} \\
    \bottomrule
  \end{tabular}
\end{table}

\begin{figure}[t]
    \centering
    \includegraphics[trim=0 5 5 5,clip,width=0.65\linewidth]{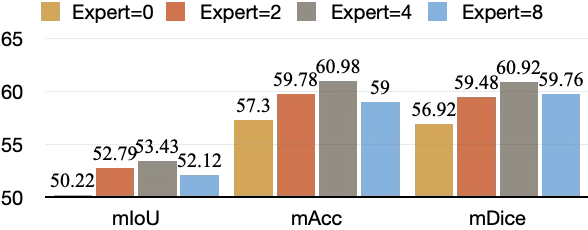}
    \caption{Effect of the number of MoE routing experts (\%).}
    \Description{Bar chart comparing mIoU, mAcc, and mDice for models using zero, two, four, or eight routing experts.}
    \label{fig:ablation_moe_expert}
\end{figure}

\noindent\textbf{Number of MoE Routing Experts.} To study the effect of routing capacity, Fig.~\ref{fig:ablation_moe_expert} varies the number of experts. Performance improves from 50.22\% mIoU without routing to 52.79\% with 2~experts and peaks at 53.43\% with 4. Increasing to 8~experts reduces mIoU by 1.31\%, indicating that an excessively large expert pool fragments the training signal and weakens specialization.

\subsection{Results on FoodSeg103}
FoodSeg103~\cite{wu2021foodseg103} is a public ingredient-level benchmark with 7,118 images across 103 categories (4,983/2,135 train/test), featuring dense spatial co-occurrence and frequent boundary interactions.

To assess cross-dataset generalization, Table~\ref{tab:foodseg_results} compares FEAST with existing methods on FoodSeg103. With ResNet-50, FEAST surpasses Mask2Former by 1.47\% mIoU, 2.99\% mDice, and 0.88\% mAcc. With Swin-B, it exceeds FDSNet by 2.92\% mIoU, 4.71\% mDice, and 1.94\% mAcc. The consistent improvements across backbones confirm that RL-guided MoE decoding generalizes beyond DishSeg24k. Additional details and ablations are provided in the supplementary material.
\begin{table}[t!]
\centering
\caption{Comparison on \textbf{FoodSeg103}~\cite{wu2021foodseg103} (\%).}
\label{tab:foodseg_results}
\renewcommand{\arraystretch}{0.80}
\resizebox{0.90\linewidth}{!}{
\begin{tabular}{l|c|ccc}
\toprule
Method & Backbone & mIoU & mDice & mAcc \\
\midrule
\multicolumn{5}{l}{\textit{CNN-based}} \\
FPN~\cite{lin2017fpn} & ResNet-50 & 27.80 & 43.51 & 38.20\\
DeepLabV3+~\cite{chen2018deeplabv3plus} & ResNet-50 & 27.81 & 39.13 & 36.63 \\ 
CCNet~\cite{huang2019ccnet} & ResNet-50 & 35.50 & 40.40 & 45.30 \\
\midrule
\multicolumn{5}{l}{\textit{Per-pixel Transformer}} \\
SegFormer~\cite{xie2021segformer} & ResNet-50 & 32.38 & 44.31 & 43.54 \\
ED-AFormer~\cite{yu2024embeddingfree} & ResNet-50 & 36.17 & 48.24 & 47.51 \\
CGRSeg~\cite{ni2024contextguided} & ResNet-50 & 31.01 & 42.54 & 41.31 \\
\midrule  
\multicolumn{5}{l}{\textit{Query-based Mask Classification}} \\
MaskFormer~\cite{cheng2021maskformer} & ResNet-50 & 35.52 & 47.71 & 47.92 \\
Mask2Former~\cite{cheng2022mask2former} & ResNet-50 & 36.38 & 48.12 & 49.43 \\
OneFormer~\cite{jain2023oneformer} & ResNet-50 & 35.16 & 46.42 & 47.73 \\
FeedFormer~\cite{shim2023feedformer} & ResNet-50 & 32.33 & 44.19 & 42.68 \\
VMFormer~\cite{yan2024vmformer} & ResNet-50 & 35.61 & 47.33 & 49.10 \\ \midrule
 
\multicolumn{5}{l}{\textit{Food-specific Methods}} \\
CCNet~\cite{wu2021foodseg103} & ReLeM-ResNet50 & 36.80 & 41.20 & 47.40\\
FoodSAM~\cite{lan2023foodsam} & ViT-H & 46.48 & 63.46 & 58.27 \\
FDSNet~\cite{Xiao2025fdsnet} & ViT-H  & 46.38 & 63.37 & 58.17 \\
FDSNet~\cite{Xiao2025fdsnet} & Swin-B & 47.34 & 64.26 & 60.04 \\ \midrule
\textbf{FEAST (Ours)} & ResNet-50 & \textbf{37.85} & \textbf{51.11} & \textbf{50.31} \\
\textbf{FEAST (Ours)} & Swin-B & \textbf{50.26} & \textbf{68.97} & \textbf{61.98} \\
\bottomrule
\end{tabular}}
\end{table}

\section{Conclusion}
We study food segmentation in real-world dining scenes with severe ambiguity, overlap, and long-tail distributions. We introduce DishSeg24k to capture these challenges and propose an RL-guided MoE decoding framework for adaptive mask refinement beyond deterministic query optimization. Beyond its strong empirical performance, FEAST shows that handling semantic entanglement requires both stronger representations and dynamic inference. DishSeg24k's fine-grained ingredient annotations and precise masks also support downstream tasks requiring accurate segmentation, including portion and volume estimation, nutritional analysis, and dietary assessment. We hope this work advances unified frameworks for comprehensive food understanding in real-world scenarios.

\begin{acks}
This work was supported by the Beijing Natural Science Foundation (JQ24021) and the National Natural Science Foundation of China (62472411 and 62125207).
\end{acks}

\bibliographystyle{ACM-Reference-Format}
\balance
\bibliography{ref}

\end{document}